\documentclass[sigconf, screen]{acmart} 

\AtBeginDocument{%
  }

\setcopyright{acmlicensed}
\copyrightyear{2024}
\acmYear{2024}

\acmSubmissionID{849}

\usepackage{amssymb}
\usepackage{multirow}
\usepackage{tabularray}
\usepackage{array}
\usepackage{colortbl}
\usepackage{booktabs}
\usepackage{caption}
\usepackage{graphicx}
\usepackage{subcaption}
\usepackage{hyperref}
\usepackage{amsfonts}
\usepackage{url}
\usepackage{makecell}
\usepackage{color}
\usepackage{xspace}
\usepackage{amsmath}
\usepackage{adjustbox}
\usepackage{booktabs}
\usepackage{algorithm}
\usepackage{algorithmic,esint}
\usepackage{soul}
\usepackage{float}
\usepackage{color}
\usepackage{enumitem}
\usepackage[normalem]{ulem}
\usepackage{tabularx}

\newcommand{\model}{\textbf{RealTCD} }
\newcommand{\modelnosp}{\textbf{RealTCD}}
\newcommand{\modelv}{RealTCD }
\newcommand{\modelvnosp}{RealTCD}

\newcommand{\blue}[1]{\textcolor{blue}{#1}}

\newcommand{\todo}[1]{%
    \ifnum#1>0%
        \blue{todo}, \todo{\numexpr#1-1\relax}%
    \fi%
}

\newcommand{\ms}[2]{{#1}\scriptsize{} $\pm$ {#2}}
\newcommand{\msone}[2]{\bf {#1}\scriptsize{} $\pm$ {#2}}

\definecolor{verylightgray}{gray}{0.1}  
\definecolor{lightblue}{RGB}{220, 240, 255}  

\begin{document}

\title{RealTCD: Temporal Causal Discovery from Interventional Data with Large Language Model}


\author{Peiwen Li}
\authornote{The work was done during author's internship at Alibaba Cloud.}
\affiliation{%
  \institution{\makebox[0pt]{SIGS, Tsinghua University}}
  \country{}
  }
\email{lpw22@mails.tsinghua.edu.cn}

\author{Xin Wang}
\authornote{Corresponding authors.}
\affiliation{
  \institution{DCST, BNRist, Tsinghua University}
  \country{}
  }
\email{xin_wang@tsinghua.edu.cn}

\author{Zeyang Zhang}
\affiliation{
  \institution{DCST, Tsinghua University}
  \country{}
  }
\email{zy-zhang20@mails.tsinghua.edu.cn}

\author{Yuan Meng}
\authornotemark[2]
\affiliation{
  \institution{DCST, Tsinghua University}
  \country{}
  }
\email{yuanmeng@tsinghua.edu.cn}

\author{Fang Shen}
\affiliation{
  \institution{Alibaba Cloud}
  \country{}
  }
\email{ziru.sf@alibaba-inc.com}

\author{Yue Li}
\affiliation{%
  \institution{Alibaba Cloud}
  \country{}
  }
\email{yueqian.ly@alibaba-inc.com}

\author{Jialong Wang}
\authornotemark[2]
\affiliation{
  \institution{Alibaba Cloud}
  \country{}
  }
\email{quming.wjl@alibaba-inc.com}

\author{Yang Li}
\affiliation{
  \institution{SIGS, Tsinghua University}
  \country{}
  }
\email{yangli@sz.tsinghua.edu.cn}

\author{Wenwu Zhu}
\authornotemark[2]
\affiliation{
  \institution{DCST, BNRist, Tsinghua University}
  \country{}
  }
\email{wwzhu@tsinghua.edu.cn}

\renewcommand{\shortauthors}{Peiwen Li et al.}

\begin{abstract}
In the field of Artificial Intelligence for Information Technology Operations, causal discovery is pivotal for operation and maintenance of systems, facilitating downstream industrial tasks such as root cause analysis.
Temporal causal discovery, as an emerging method, aims to identify temporal causal relations between variables directly from observations by utilizing interventional data. 
However, existing methods mainly focus on synthetic datasets with heavy reliance on interventional targets and ignore the textual information hidden in real-world systems, failing to conduct causal discovery for real industrial scenarios. 
To tackle this problem, in this paper we investigate temporal causal discovery in industrial scenarios, which faces two critical challenges: how to discover causal relations without the interventional targets that are costly to obtain in practice, and how to discover causal relations via leveraging the textual information in systems which can be complex yet abundant in industrial contexts.
To address these challenges, we propose the \model framework, which is able to leverage domain knowledge to discover temporal causal relations without interventional targets. 
We first develop a score-based temporal causal discovery method capable of discovering causal relations without relying on interventional targets through strategic masking and regularization.
Then, by employing Large Language Models (LLMs) to handle texts and integrate domain knowledge, we introduce LLM-guided meta-initialization to extract the meta-knowledge from textual information hidden in systems to boost the quality of discovery. 
We conduct extensive experiments on both simulation datasets and our real-world application scenario to show the superiority of our proposed \model over existing baselines in temporal causal discovery.
\end{abstract}




\maketitle

\section{Introduction}
The advent of Artificial Intelligence for Information Technology Operations (AIOps) has revolutionized the way we manage and operate complex information systems. Causal discovery plays a pivotal role in understanding the intricate network of dependencies and influences within these systems~\cite{mogensen2023causal, arya2021evaluation, zhang2023outofdistribution,zhang2021revisiting}, offering invaluable insights for various downstream industrial tasks in AIOps, including anomaly detection~\cite{yang2022causality} and root cause analysis~\cite{yang2022causal, wang2023incremental} etc. For instance, by equipping AIOps with the ability to accurately identify the underlying causal structures, AIOps systems can effectively detect abnormal behaviors and determine the underlying causes of system failures, thus leading to enhanced operational efficiency and improved decision-making processes in industry. 

Temporal causal discovery, as an emerging approach, aims to directly identify temporal causal relationships between variables based on observational data, with the utilization of interventional data. This group of methods has gained significant attention in recent years due to their promising potential to uncover causal dependencies in dynamic systems. 
\citet{brouillard2020differentiable} and \citet{li2023causal} employ temporal causal discovery methods to leverage various types of interventional data and have achieved remarkable progress in discovering the underlying temporal causal relationships. 

However, the existing studies mainly focus on studying synthetic datasets, which strongly rely on interventional targets and ignore the intricate complexities and nuances hidden in real-world systems, failing to conduct causal discovery for real industrial scenarios. In this paper, we tackle this problem by studying temporal causal discovery in industrial scenarios, which is non-trivial and poses the following two critical challenges: 

\begin{itemize}[leftmargin = 0.5cm]
    \item How to discover casual relationships without the interventional targets that are normally costly to obtain in practice?
    \item How to discover causal relationships via leveraging the textual information in systems which can be complex yet abundant in industrial contexts?
\end{itemize}

To address these challenges, we propose the \model framework, which is able to leverage the textual information from real-world systems to discover temporal causal relationships without interventional targets. 
Specifically, we first develop a score-based temporal causal discovery method that learns the underlying causal relationship without interventional targets through strategic masking and regularization. 
We impose regularizations on both the adjacency matrix and interventional family within the context of the regularized maximum log-likelihood score, and optimize them in a joint manner. In this way, the costly interventional targets are not required for boarder applications in real-world industrial scenarios. Subsequently, by leveraging Large Language Models (LLMs) to handle texts, we introduce LLM-guided meta-initialization that infers and initializes the inherent causal structures from the textual information in systems for the aforementioned discovery process, which incorporates the domain knowledge while upholding the theoretical integrity of temporal causal discovery. Extensive experiments on both simulation and real-world datasets demonstrate the superiority of our \model framework over existing baselines.
Deeper analyses also show that our method can effectively discover the underlying temporal causal relationships without interventional targets in industrial scenarios. In summary, our main contributions are as follows:

\begin{itemize}[leftmargin=0.5cm]
    \item We study the problem of temporal causal discovery in industrial scenarios. To the best of our knowledge, we are the first to solve the problem with Large Language Models (LLMs) and without interventional targets.
    \item We propose the \model framework, including two specially designed modules: score-based temporal causal discovery and LLM-guided meta-initialization, which is able to leverage the textual information in systems to discover temporal causal relationships without interventional targets in industrial scenarios.
    \item Extensive experiments on both simulation and real-world datasets demonstrate the superiority of our framework over several baselines in discovering temporal causal structures without interventional targets.
\end{itemize}

\section{Preliminary}

\subsection{Dynamic Causal Graphical Model}\label{Sec:DCGM}

Since causal graphical models (CGMs) support interventions compared with standard Bayesian Networks, we introduce the dynamic causal graphical models (DyCGMs) extended from CGMs in order to formulate interventions between variables across time slices. Suppose that there are $d$ different measuring points in a system, and we consider causality within $p$ time-lagged terms. Therefore, the object of our study on temporal causal discovery is actually $(p+1)\times d$ random variables $N_{0,1}, \ldots, N_{0,d},\ldots, N_{p,1}, \ldots, N_{p,d}$, where $N_{k,l}, k\in \{0,\ldots,p\}, l\in \{1,\ldots,d\}$ denotes the $k$ time-lagged version of the $l$th measuring point. For the convenience of subsequent presentation, we abbreviate the above variables in order as $X_{i},i\in \{1,\ldots,(p+1)d\}$. 

Based on this, a DyCGM is defined by the distribution $P_X$ over the vector $X=(X_{1}, \ldots, X_{(p+1)d})$ and a DAG $\mathcal{G}=(V, E)$. 
To be specific, each node $i\in V=\{1,\ldots,(p+1)d\}$ is related with a random variable $X_i,i\in \{1,\ldots,(p+1)d\}$, and each edge $(i,j)\in E$ represents a direct causal relation from variable $X_i$ to $X_j$.
Under the Markov assumption of the distribution $P_Y$ and graph $\mathcal{G}$, the joint distribution can be factorized as 
\begin{equation}
p(x_1,\ldots,x_{(p+1)d}) = \prod \limits_{j=1}^{(p+1)d} p_j (x_{j} |x_{\pi_j^\mathcal{G}} ),
\end{equation}
where $\pi_j^\mathcal{G}$ is the set of parents of the node $j$ in the graph $\mathcal{G}$, and $x_{\pi_j^\mathcal{G}}$ denotes the entries of the vector $x$ with indices in $\pi_j^\mathcal{G}$. 

We also assume \emph{causal sufficiency}, which means there is no hidden common cause that is causing more than one variable in $X$~\cite{pearl2009causality}.

\subsection{Intervention} \label{Sec:intervention}
An intervention on a variable $x_{j}$ corresponds to replacing its conditional density $p_j (x_{j} |x_{\pi_j^\mathcal{G}})$ by a new one. Apart from that, we define the \emph{\textbf{interventional target}}, a set $I\subseteq V$ consisting of the variables being intervened simultaneously, and the \emph{\textbf{interventional family}} $\mathcal{I}:=(I_1, \ldots, I_Q)$, where Q is the number of interventions. To be specific, the observational setting, where no variables were intervened, is always known and denoted by $I_1:=\emptyset$. The $q$th interventional joint density can be represented as
\begin{equation}\label{interv}
p^{(q)}(x_1,\ldots,x_{(p+1)d}) := 
\prod \limits_{j\not\in I_q} p_j^{(1)}(x_{j} |x_{\pi_j^\mathcal{G}})
\prod \limits_{j\in I_q} p_j^{(q)}(x_{j} |x_{\pi_j^\mathcal{G}}).
\end{equation}

Note that, in the temporal domain, merely the contemporary variables $N_{0,1}, \ldots, N_{0,d}$, i.e. $X_1, \ldots, X_d$ can be intervened and be in an interventional target, as only time-lagged variables $N_{k,l}, k\in \{1,\ldots,p\}, l\in \{1,\ldots,d\}$, i.e. $X_i, i\in \{d+1,\ldots,(p+1)d\}$ and other contemporary variables $N_{0,l}, l\in \{1,\ldots,d\}\setminus \{j\}$, i.e. $X_i, i\in \{0,\ldots,d\}\setminus \{j\}$ can affect a particular contemporary variable $N_{0,j}$, i.e. $X_j$.

Meanwhile, there are two types of interventions: 1) imperfect (or soft, parametric) interventions is the general type depicted below, and 2) perfect interventions (or hard, structural)~\cite{jaber2020causal} is a special case that removes the dependencies of an intervened node $j\in \mathcal{I}_q$ on its parent nodes, i.e. $p_j^{(q)}(x_{j} |x_{\pi_j^\mathcal{G}}) = p_j^{(q)}(x_{j})$ in equation~(\ref{interv}).

\section{Method}
\begin{figure*}[h]
  \centering
  \includegraphics[width=0.9\linewidth]{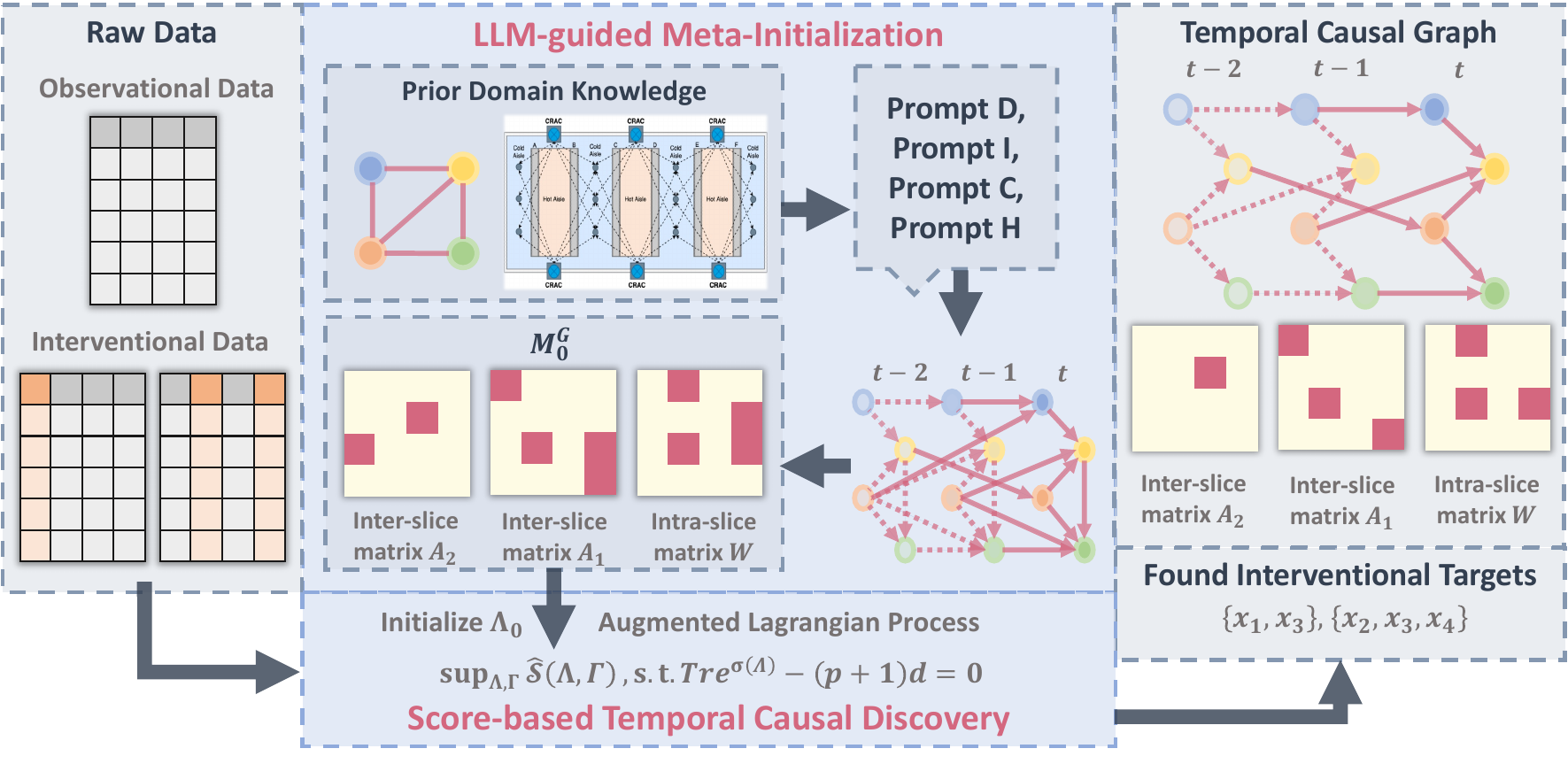}
  \caption{
  The framework of our proposed method \modelnosp. Given the system textual information and temporal data without interventional targets, the LLM-guided Meta-Initialization module leverages LLMs to extract the domain knowledge and obtain the potential causal relationships as the initialization adjacency matrix $M_0^\mathcal{G}$. Then, the Score-based Temporal Causal Discovery module utilizes an augmented Lagrangian process to optimize the score for unknown interventional targets under constraints, where the $\Lambda_0$ is initialized with $M_0^\mathcal{G}$. In this way, the proposed \model leverages the system textual information to discover temporal causal relationships without interventional targets.
  }
\label{fig:framework}
\end{figure*}

In order to make the causal discovery process able to be applied to the real industrial scene effectively, we propose the \model framework as shown in Figure~\ref{fig:framework}, including two modules: 1) in the module \textbf{Score-based Temporal Causal Discovery}, data from normal state and abnormal state of a system under AIOps are modeled as observational data and interventional data respectively, and relaxation is done to the condition that the label of interventional targets is known, making the algorithm easier to apply to real scenes; 2) in the module \textbf{LLM-guided Meta Initialization}, LLM is leveraged to introduce the domain knowledge and system structure information in text types and to preliminarily obtain possible causal relations from them as initialization for the discovery process.

\subsection{Score-based Temporal Causal Discovery}
In this section, we introduce a score-based purely data-driven temporal causal discovery from interventional data with unknown interventional targets.

\subsubsection{Raw data}
Raw data we used as shown in Figure~\ref{fig:framework} refers to a set of input temporal data including both observational and interventional data. Observational data refers to standard data without interventions or anomalies, while interventional data contains interventions discussed in Section~\ref{Sec:intervention}, or represents abnormal data in real-world datasets. The colored columns represent time series of ground truth interventional targets that are unknown in prior. There may be multiple sets of interventional data, each corresponding to different interventional targets or anomalies.

\subsubsection{Model conditional densities}

To begin with, we use neural networks to model conditional densities.
Firstly, we encode the DAG $\mathcal{G}$ with a binary adjacency matrix $M^{\mathcal{G}}\in\{0,1\}^{(p+1)d\times (p+1)d}$ which acts as a mask on the neural network inputs.
Similarly, we encode the interventional family $\mathcal{I}$ with a binary matrix $R^\mathcal{I}\in\{0, 1\}^{Q\times (p+1)d}$, where $ R^\mathcal{I}_{qj} = 1$ means that $X_j$ is an intervened node in the interventional target set $I_q$. Then, following equation~(\ref{interv}), we further model the joint density of the $q$th intervention by

{
\begin{equation}
\label{density}
\begin{aligned}
    f^{(q)}\left(x ; M^{\mathcal{G}}, R^{\mathcal{I}}, \phi\right):=\prod_{j=1}^{(p+1)d} \tilde{f}\left(x_j ; \mathrm{NN}\left(M_j^{\mathcal{G}} \odot x ; \phi_j^{(1)}\right)\right)^{1-R_{q j}^{\mathcal{I}}} \\
\tilde{f}\left(x_j ; \mathrm{NN}\left(M_j^{\mathcal{G}} \odot x ; \phi_j^{(q)}\right)\right)^{R_{qj}^{\mathcal{I}}}
\end{aligned},
\end{equation}}
where $\phi:= \{\phi^{(1)},\ldots,\phi^{(Q)}\}$, the NN’s are neural networks parameterized by $\phi^{(1) }_j$ or $\phi^{(q)}_j $, the operator $\odot$ denotes the Hadamard product (element-wise) and $M^\mathcal{G}_j$ denotes the $j$th column of $M^\mathcal{G}$, enabling $M_j^{\mathcal{G}}\odot x$ to select the parents of node $j$ in the graph $\mathcal{G}$.

\subsubsection{Score for unknown interventional targets}
Based on the NN conditional densities in equation~(\ref{density}), we can firstly formulate the following \textbf{regularized maximum log-likelihood score} as the basic score for known interventional targets' setting:
{
\begin{equation}
\label{knownscore}
\mathcal{S}_{\mathcal{I}^*}(\mathcal{G}):=\sup_\phi \sum_{q=1}^Q \mathbb{E}_{X \sim p^{(q)}} \log f^{(q)}\left(X, M^{\mathcal{G}}, R^{\mathcal{I}^*}, \phi\right)-\lambda|\mathcal{G}|,
\end{equation}}
where the ground truth interventional family (containing the interventional targets) $\mathcal{I}^*:=(I_1^*, \ldots, I_Q^*)$ is known and $p^{(q)}$ stands for the $q$th ground truth interventional distribution, $|\mathcal{G}|$ represents the number of edges in the causal graph. By maximizing the score in equation~(\ref{knownscore}), we can get an estimated DAG $\hat{\mathcal{G}}$ that is $\mathcal{I}^*$-Markov equivalent to the true DAG $\mathcal{G}^*$~\cite{brouillard2020differentiable}, under the condition that the ground truth interventional family is known.

Then, we assume the interventional targets are unknown. To still be able to utilize the special information from interventional data, we propose to jointly optimize the adjacency matrix and interventional family as well as the NN's parameters, thus, reaching the two optimization goals simultaneously. To be specific, a regularization term for the interventional family is added to the above score, and we form score for unknown interventional targets:
\begin{equation}
\label{unknownscore}
\mathcal{S}(\mathcal{G},\mathcal{I}):=\sup_\phi \sum_{q=1}^Q \mathbb{E}_{X \sim p^{(q)}} \log f^{(q)}\left(X, M^{\mathcal{G}},R^{\mathcal{I}}, \phi\right)-\lambda|\mathcal{G}|-\lambda_R|\mathcal{I}|,
\end{equation}
where $|\mathcal{I}|=\sum_{q=1}^Q |\mathcal{I}_q|$ counts the total number of intervened nodes.

The following theorem guarantees the identification of the temporal causal graph as well as the interventional family under the setting that interventional targets are unknown, which can be proved similarly as in our previous work~\cite{li2023causal}.
\begin{theorem}[Unknown targets temporal causal DAG identification]
    Suppose $\mathcal{I}^*$ is such that $\mathcal{I}_1^*:= \emptyset$. Let $\mathcal{G}^*$ be the ground truth temporal DAG and $(\hat{\mathcal{G}},\hat{\mathcal{I}})\in arg max_{\mathcal{G}\in DAG,\mathcal{I}}\mathcal{S}(\mathcal{G},\mathcal{I})$. 
    Under the assumptions that: 1) the density model has enough capacity to represent the ground truth distributions; 2) $\mathcal{I}^*$-faithfulness holds; 3) the density model is strictly positive; 4) the ground truth densities $p^{(q)}$ have finite differential entropy.
    For $\lambda, \lambda_R>0$ small enough, $\hat{\mathcal{G}}$ is $\mathcal{I}^*-Markov$ equivalent to $\mathcal{G}^*$ and $\hat{\mathcal{I}}=\mathcal{I}^*$.
\end{theorem}

\subsubsection{Maximize the score}

Subsequently, to allow the gradient-based stochastic optimization process, we relax the above score by taking $M^{\mathcal{G}}$ and $R^\mathcal{I}$ as a random matrix respectively, where $M_{ij}^{\mathcal{G}} \sim B(1,\sigma(\alpha_{ij}))$ and $R_{qj}^{\mathcal{I}} \sim B(1,\sigma(\beta_{qj}))$, $B$ represents the Bernoulli distribution, $\sigma$ is the sigmoid function and $\alpha_{ij}, \beta_{qj}$ are scalar parameters. We group these $\alpha_{ij}$s into a matrix $\Lambda \in \mathbb{R}^{(p+1)d\times (p+1)d}$, and $\beta_{kj}$s into a matrix $\Gamma \in \mathbb{R}^{Q\times (p+1)d}$.
After that, we rely on \textit{augmented Lagrangian procedure}~\cite{zheng2018dags} to maximize the following score:
{
\begin{multline}
\label{finalscore}
\hat{\mathcal{S}}(\Lambda,\Gamma):=\sup _\phi \underset{M \sim \sigma(\Lambda)}{\mathbb{E}} \\
\left[ \underset{R \sim \sigma(\Gamma)}{\mathbb{E}} \left[\sum_{q=1}^Q \underset{X \sim p^{(q)}}{\mathbb{E}} \log f^{(q)}\left(X ; M, R^{\mathcal{I}^*}, \phi\right) 
-\lambda\|M\|_0 -\lambda _R \|R\|_0 \right] \right],
\end{multline}}
under the acyclicity constraint:
{
\begin{equation}
\label{constrain}
\sup_{\Lambda,\Gamma} \hat{\mathcal{S}}(\Lambda,\Gamma),
\text{s.t.}\operatorname{Tr}e^{\sigma(\Lambda)}-(p+1)d=0.
\end{equation}}

Moreover, as for gradient of the score w.r.t. $\alpha_{ij}$ and $\beta_{qj}$, following the general dealing method in continuous optimization for causal discovery~\cite{kalainathan2022structural, ng2022masked}, we estimate $\Lambda$ and $\Gamma$ by \textit{Straight-Through Gumbel estimator}, which means that Bernoulli samples are used in forward pass and Gumbel-Softmax samples are used in backward pass.

Overall, the learnable parameters in the process are $\phi$,$\Lambda$, $\Gamma$, and the estimated adjacency matrix reflecting temporal causal relations is $\sigma(\Lambda)$ and the estimated potential interventional family is $\sigma(\Gamma)$.

Since we only focus on influences on $X_1,\ldots, X_d$ from other variables, we set $\Lambda[:,d+1:(p+1)d]$, i.e. the meaningless part $M^\mathcal{G}[:,d+1:(p+1)d]$, to zero before training.

\subsection{LLM-guided Meta Initialization}

In this section, we introduce the detailed method of using LLM to bring in domain knowledge and extra prior information~\cite{castelnovo2024marrying,pawlowski2023answering} so that the potential temporal causal relations are obtained to guide the data-driven optimization process. 

\paragraph{Prompts of LLMs} We describe the system into prompts as queries to the LLMs for possible temporal causal relationships. One prompt example is shown in Table~\ref{tab:prompt}, where the underlined units are indispensable and the others are optional. 
We also provide the more detailed example of the prompt used for the real-world dataset in our experiments in Appendix~\ref{App:prompts}.

We have implemented strategies to mitigate such biases, notably through the use of tailored prompts (Prompt D and Prompt I). Prompt D is designed to clarify the causal discovery tasks for the LLMs and align them with the domain knowledge inherent to the LLMs themselves. Prompt I further introduces data and domain knowledge consistent with the subsequent causal discovery tasks to the LLMs, such as context descriptions, physical structures, and generating rules, to ensure that the meta-initialization process is as unbiased as possible.

Using tuples provided by LLM, we construct the adjacency matrix $M_0^\mathcal{G}$ to denote the learned causal structure. These results represent potential causality based on domain information, not guaranteed causal definitions. \textit{The subsequent score-based causal discovery optimization ensures the final results conform to causal definitions.} We incorporate meta-initialization information as follows:
  

\begin{table}
\small
\centering
\setlength{\extrarowheight}{0pt}
\addtolength{\extrarowheight}{\aboverulesep}
\addtolength{\extrarowheight}{\belowrulesep}
\setlength{\aboverulesep}{0pt}
\setlength{\belowrulesep}{0pt}
\caption{Example prompt for LLM-guided Meta Initialization. }
\label{tab:prompt}
\vspace{-1pt}
\begin{tabular}{>{\hspace{0pt}}m{0.95\linewidth}} 
\toprule
\rowcolor{lightblue} \textbf{Prompt D}efinition                         \\
\underline{\texttt {\textbf{Role}}}: 
"You are an exceptional temporal causal discovery analyzer, with in-depth domain knowledge in \ldots (e.g. the intelligent operation and maintenance of data center air-conditioning systems)."

\underline{\texttt {\textbf{Introduction}}}: 
"A directed temporal causal relationship between variables xu and xv can be represented as a tuple (xu, xv, t), signifying that the variable xu, lagging t time units, causally influences the current state of variable xv. The tuple (xu, xv, 0) denotes contemporaneous causality if t=0; if t>0, the tuple (xu, xv, t) indicates time-lagged causality. Note that when t=0, i.e. in (xu, xv, 0), xu and xv must be different variables, as intra-slice self-causality is not considered. Also, (xu, xv, 0) and (xu, xv, t) for t>0 have the possibility to coexist, suggesting that contemporaneous and time-lagged causality between two variables might simultaneously occur sometimes.
Our task is to unearth all the possible temporal causal relationships among variables, grounded on the subsequent information."           \\ 
\noalign{\kern-\cmidrulewidth}\cmidrule{1-1}
\rowcolor{lightblue} \textbf{Prompt I}nformation                        \\
\underline{\texttt{\textbf{Domain knowledge}}}: 
Depending on the application scenarios, it might be: 1) A \textit{context description} of a specific industrial scenario or AIOps scenario. 2) The \textit{physical structure} of the system, containing the \textit{location} information of each entity or variable. 3) The abstract \textit{generating rules} of time series.

\texttt{\textbf{Data}}: 
Providing a piece of past time series may help LLM understand the variables and their relations better.                                                                                        \\ 
\noalign{\kern-\cmidrulewidth}\cmidrule{1-1}
\rowcolor{lightblue} \textbf{Prompt C}ausal Discovery in Temporal Domain  \\
\underline{\texttt{\textbf{Task}}}: 
"Please identify all temporal causal relations among the $n$ variables ($x_1, \ldots, x_n$), considering only contemporaneous and $p$ time-lagged causality. Conclude your response with the full answer as a Python list of tuples (xu, xv, t) after 'Answer:'. Don't simplify and just give me some examples. You should cover all possible relationships in your answer."
\\ 
\noalign{\kern-\cmidrulewidth}\cmidrule{1-1}
\rowcolor{lightblue} \textbf{Prompt H}int                               \\
\texttt{\textbf{Implication}}: 
We can offer a thinking path to the LLM model as a guide of how we want it to utilize the prior information we gave to it or its own knowledge and deduct the answer.

\texttt{\textbf{Chain of Thought (CoT)}}: 
"Proceed methodically, step by step." By simply adding a zero-shot CoT prompt, the LLM model could output its answer with its path of thought, making the process more interpretable and easy for humans to understand, check, and correct immediately~\cite{wei2022chain,lyu2023faithful}. Furthermore, if we can provide an example that contains the correct thought path and result as input, the one-shot CoT may achieve an even better result.\\
\bottomrule
\end{tabular}
\vspace{-10pt}
\end{table}

\paragraph{Guide temporal causal discovery from data}
By initiating $M^\mathcal{G}$ or to say $\Lambda$ of the module score-based temporal causal discovery as the results $M_0^\mathcal{G}$ from the module LLM-guided meta initialization  before training and optimization, we not only lead the direction of the data-driven optimization process but also guarantee the theoretical integrity of the temporal causal discovery from interventional data.

\paragraph{Extract weight}
Eventually, we obtain the estimated full weight matrix $\hat{F}=\sigma(\Lambda)$ of graph $\mathcal{G}$. Then, we form the adjacency matrix $\hat{M}^\mathcal{G}$ of the graph by adding an edge whenever $\sigma(\Lambda)>0.5$ is acyclic.
Finally, we can extract the intra-slice matrix $\hat{W}=\hat{M}^\mathcal{G} [1:d,1:d]$ and inter-slice matrix $\hat{A}_k=\hat{M}^\mathcal{G} [kd+1:(k+1)d,1:d]$ for each time lag $k=1,\ldots,p$. They reflect causal relations of these $d$ variables in both a contemporary and time-lagged manner.

The overall framework and algorithm of \model are summarized in Figure~\ref{fig:framework} and Algorithm~\ref{tab:pipeline1} respectively.      
\begin{algorithm}
\small
\caption{The overall algorithm for \model} 
\label{tab:pipeline1}
\begin{algorithmic}[1]
\REQUIRE All kinds of prior knowledge
\STATE Generate prompt as described in Table~\ref{tab:prompt}
\STATE Obtain causal results from LLM and transfer into matrix $M_0^\mathcal{G}$
\REQUIRE ~$M_0^\mathcal{G}$, hyperparameter $\lambda$, $\lambda_R$, hyperparameter for augmented Lagrangian process
\STATE Transfer the constrained problem defined by equation~(\ref{finalscore}), (\ref{constrain}) into the form of unconstrained problem following augmented Lagrangian
\STATE Initialize $\Lambda_0$ by $M_0^\mathcal{G}$, and also initialize $\phi_0, \Gamma_0$, max\_iteration $U$, Lagrangian multiplier $\gamma_0$ and penalty coefficient $\mu_0$
\WHILE{$0\leq t\leq U$ and $h(\Lambda):=\operatorname{Tr}e^{\sigma(\Lambda)}-(p+1)d>10^{-8}$}
\STATE Solve the $t$th unconstrained subproblem using stochastic gradient descent algorithm (we use RMSprop)
\STATE Get sub solution $\phi_t^*,\Lambda_t^*,\Gamma_t^*$, and initialize $\phi_{t+1},\Lambda_{t+1},\Gamma_{t+1}$ by them
\STATE Update~$\gamma_{t+1}$ and $\mu_{t+1}$
\STATE $t=t+1$
\ENDWHILE
\STATE Form causal graph by adding an edge whenever $\sigma(\Lambda)>0.5$ is acyclic
\end{algorithmic}
\end{algorithm}

\section{Experiments}
\subsection{Setups}
\subsubsection{Baselines}
To evaluate the effectiveness of our method, we compare with the following models as baselines:
\begin{itemize}[leftmargin=0.5cm]
    \item \textbf{DYNOTEARS}~\cite{pamfil2020dynotears}: As DYNOTEARS is a method focused on fitting the exact values of time series, it outputs quantitative weight values. We set threshold value of $W$ and $A$ to be $0.5$.
    \item \textbf{PCMCI}~\cite{runge2020discovering}: We use results with a significance level of 0.01. 
    \item \textbf{TECDI}~\cite{li2023causal}: It is a temporal causal discovery methods require both interventional data and interventional targets. 
    \item \textbf{NeuralGC}~\cite{tank2021neural}: Since NeuralGC only learns contemporaneous relationships, we use $W_{\mathrm{full}}$ defined as below to compress both contemporaneous and time-lagged causal relationships learnt from \modelv into a single metric for comparison with NeuralGC, which does not differentiate these 2 types. {
    \begin{equation}\label{eq:Wfull}
    W_{\mathrm{full}}(i,j) = 
    \begin{cases}
    1, & W(i,j)+\sum_{k=1}^p A_{k}(i,j) > 0 \\
    0, & \text{otherwise}
    \end{cases}
    \end{equation}}
    This formulation captures the entire causal relationship from $i$ to $j$, assuming a relationship exists if any contemporaneous or time-lagged relationship is present.
    \end{itemize}

The distinction in method selection across different settings was intentional and aligned with the capabilities of each algorithm: Since PCMCI and DYNOTEARS are designed for causal structure learning from observational data and can not incorporate interventional data, we  included them in the "Unknown" interventional targets setting. 
As for TECDI, we use the same data with \modelvnosp.
For baselines that are unable to bring in interventional data, we ensure a fair comparison by keeping the overall sample size consistent across all models, while using only observational data (or normal data in real datasets) for those baselines.

\subsubsection{Synthetic datasets}

We generate temporal data in two steps: 
\begin{itemize}[leftmargin=0.5cm]
    \item Sample intra DAG and inter DAG following the \textit{Erdős-Rényi} scheme, then sample parameters in weighted adjacency matrix, where elements in intra-slice matrix $W$ are uniformly from $[-1.0,-0.25]\cup[0.25,1.0]$ and  elements in inter-slice matrixes $A_k$ are uniformly from $[-1.0\alpha,-0.25\alpha]\cup[0.25\alpha,1.0\alpha],  
    \alpha=1/\eta^{k}, \eta\geq1, k=1,\ldots,p$. 
    \item Generate time series consistent with the sampled weighted graph following the standard structural vector autoregressive (SVAR) model\cite{rubio2010structural}: $Y_0= Y_0 W+Y_1 A_1+\cdots+Y_p A_p+Z$, where $Z$ is random variables under the normal distribution.
Then, sample interventional targets from nodes in $Y_0$, and generate  perfect interventional data by cutting off the dependency of intervened nodes on their parents, i.e. setting $W_{ij}$ and ${A_k}_{ij}$ to zero, where $x_j$ is the variable in interventional targets and $x_i\in x_{\pi_j^\mathcal{G}}$.
\end{itemize}

Before training, all data are normalized by subtracting the mean and dividing by the standard deviation.
We experimented on two simulated datasets: Dataset 1 contains 5 nodes, their 1 time-lagged variables and 5 different interventional targets.
Dataset 2 contains 10 nodes, their 1 time-lagged variables and 10 different interventional targets.
We initially chose to limit our datasets to a time delay of 1 to facilitate the evaluation and presentation. However, it is indeed feasible to increase the time delay of the data.

\subsubsection{Real-world data center application}

\begin{figure}[h]
  \centering
  \includegraphics[width=0.9\linewidth]{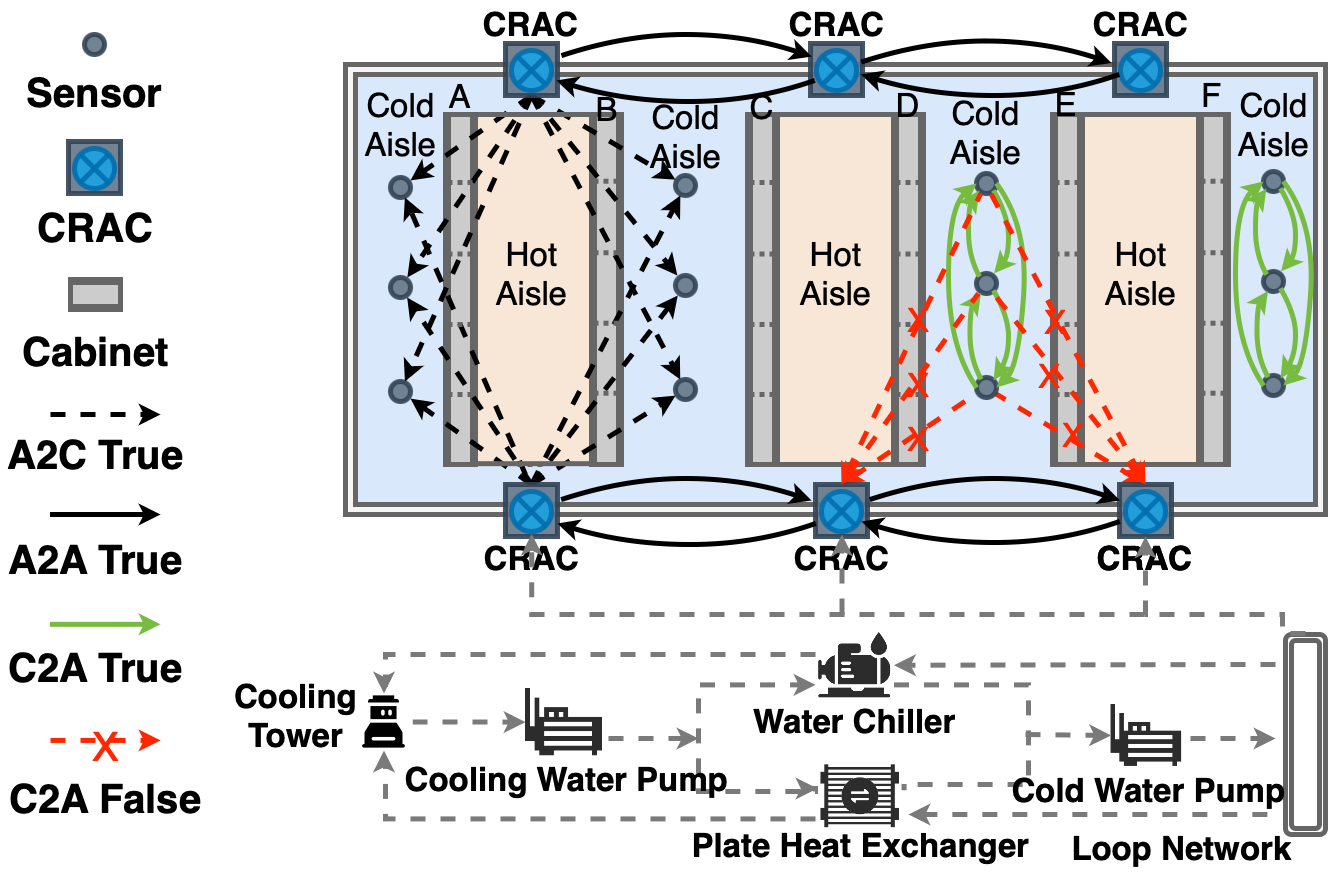}
  \caption{A typical data center cooling system diagram.}
  \label{fig:realgraph}
\end{figure}
In contemporary data centers, IT equipment stability is crucial. Sophisticated air conditioning systems manage heat, maintaining a consistent temperature. Figure~\ref{fig:realgraph} shows a typical data center room organized with rows of equipment (A, B, C, D, etc.), separated by barriers to prevent hot and cold air mixing. Computer Room Air Conditioners (CRACs) on both sides of the room create a closed-loop system to maintain stable conditions. Multiple sensors in the cold aisle provide real-time temperature data, essential for ensuring continuous cold air delivery and stable IT operation.


When an anomaly occurs at a monitoring point, we can find the root cause of the anomaly based on the causal relationships among these entities we have learned through our \model framework.

\paragraph{Data acquisition.}
The data used in this study was obtained from a specific data center at Alibaba. It covers monitoring data from a cooling system of a particular room from January 1st, 2023 to May 1st, 2023, and includes 38 variables in total. These variables comprise \textit{18 cold aisle temperatures} from sensors and \textit{20 air conditioning supply temperatures} from CRACs.
We collected several time series during normal as well as abnormal states. For the latter, data was sampled within 20 minutes of the occurrence of the abnormality, with each sampling interval being 10 seconds.
Anomaly points were identified by learning the normal distribution range from historical data, using the $n$-$\sigma$ method. Any data points that fall outside of the $n$-$\sigma$ range (e.g., 3 to 5) of itself are extracted as anomaly time points. 

A more detailed description of synthetic and real-world datasets generation is in Appendix~\ref{App:datasets}.

\subsubsection{Evaluation metrics}
\paragraph{For synthetic datasets}
We leverage two metrics to evaluate the performance of learning causal graph: {\romannumeral1}) structural Hamming distance (\textbf{SHD}), which calculates the number of different edges (either reversed, missing or redundant) between two DAGs; {\romannumeral2}) structural interventional distance (\textbf{SID}), which represents the difference between two DAGs according to their causal inference conditions\cite{peters2015structural}. 
\paragraph{For real-world datasets}
Given the absence of ground truth DAGs in real case, we employ four performance metrics based on expert knowledge to assess algorithms' efficacy. These metrics mainly consider the physical location relationships within the room, shown in figure~\ref{fig:realgraph}. 
{\romannumeral1}) \textbf{A2C True} counts the correctly identified causal edges from \textit{air conditioning supply temperatures} (\textit{A}) to the \textit{temperatures of} adjacent \textit{cold aisles} (\textit{C}), assuming that \textit{A} influences \textit{C}.
{\romannumeral2}) \textbf{A2A True} counts the correctly identified causal edges between adjacent \textit{A} units, assuming mutual influences among them.
{\romannumeral3}) \textbf{C2C True} counts the correctly identified causal edges between temperatures in the same column of cold aisles, assuming direct interactions among them.
{\romannumeral4}) \textbf{C2A False} counts the incorrectly identified causal edges from cold aisle temperatures (C) to air conditioning supply temperatures (A), as such causality is implausible given that downstream variables (C) cannot influence upstream variables (A).

\subsection{Main Results}


\subsubsection{On synthetic datasets}

The results on synthetic datasets are reported in Table~\ref{tab:simu}.
On both datasets, our method outperforms baseline models on SHD and SID metrics, with small standard deviations. 
For the dataset with 10 nodes, the improvement is more pronounced, highlighting our method's advantages in handling a \textbf{large number of variables}, making optimization more \textbf{focused} and \textbf{effective}. 
\modelv consistently achieves better performance in each setting. "Known" targets setting performs better than the "Unknown" targets setting since the former employs ground-truth interventional target labels for training. 
However, since ground-truth targets are often unavailable in real scenarios, \modelvnosp's effectiveness with unknown targets is highly practical.
\begin{table*}[h]
\small
\centering
\caption{Results on synthetic datasets. $\uparrow$ denotes the higher the better, and $\downarrow$ denotes the lower the better. The best results are in bold in the setting of unknown interventional targets.}
\label{tab:simu}
\begin{tabular}{lllcccc}
\toprule
                         &       &     & \multicolumn{2}{c}{5 nodes, 1 lag} & \multicolumn{2}{c}{10 nodes, 1 lag} \\
                         \cmidrule(r){ 4-5 } \cmidrule(r){6-7}
Targets& Causality& Method& SHD $\downarrow$ & SID $\downarrow$& SHD $\downarrow$ & SID $\downarrow$ \\
\midrule
\multirow{2}{*}{Known}&\multirow{2}{*}{intra+inter}& TECDI      & \ms{1.55}{2.30} & \ms{1.82}{2.40} & \ms{3.91}{5.05}   & \ms{10.64}{10.35}\\
 &  & \modelv & \msone{1.20}{2.80}&\msone{1.56}{3.12}&\msone{2.15}{4.32}&\msone{8.95}{10.54}\\
\midrule
\multirow{6}{*}{Unknown}  &\multirow{4}{*}{intra+inter}& DYNOTEARS  & \ms{20.40}{2.41}  & \ms{38.60}{3.72}  & \ms{36.00}{5.21}  & \ms{118.60}{20.66}\\
                       &  & PCMCI      & \ms{18.10}{4.43}   & \ms{24.10}{3.11}  & \ms{62.00}{17.15}  & \ms{118.30}{24.08} \\
                       &  & TECDI & \ms{11.90}{4.75}  & \ms{17.70}{8.49}&\ms{27.40}{10.69}&\ms{60.80}{25.74} \\
                       &  & \modelv & \msone{9.90}{2.85} & \msone{16.10}{6.19}&\msone{7.10}{4.43}   & \msone{18.50}{11.04}\\
                       \cmidrule(r){2-7} 
 & \multirow{2}{*}{intra} & NeuralGC & \ms{14.90}{2.28}&\ms{20.00}{0.00}&\ms{31.30}{4.72}&\ms{85.50}{6.36}\\
 &  & \modelv & \msone{6.30}{1.83}&\msone{18.10}{3.75}&\msone{5.62}{3.81}&\msone{75.25}{11.89}\\
\bottomrule
\end{tabular}
\end{table*}

\begin{figure}[h]
  \centering
  \begin{subfigure}[b]{0.9\linewidth}
    \centering
    \includegraphics[width=\linewidth]{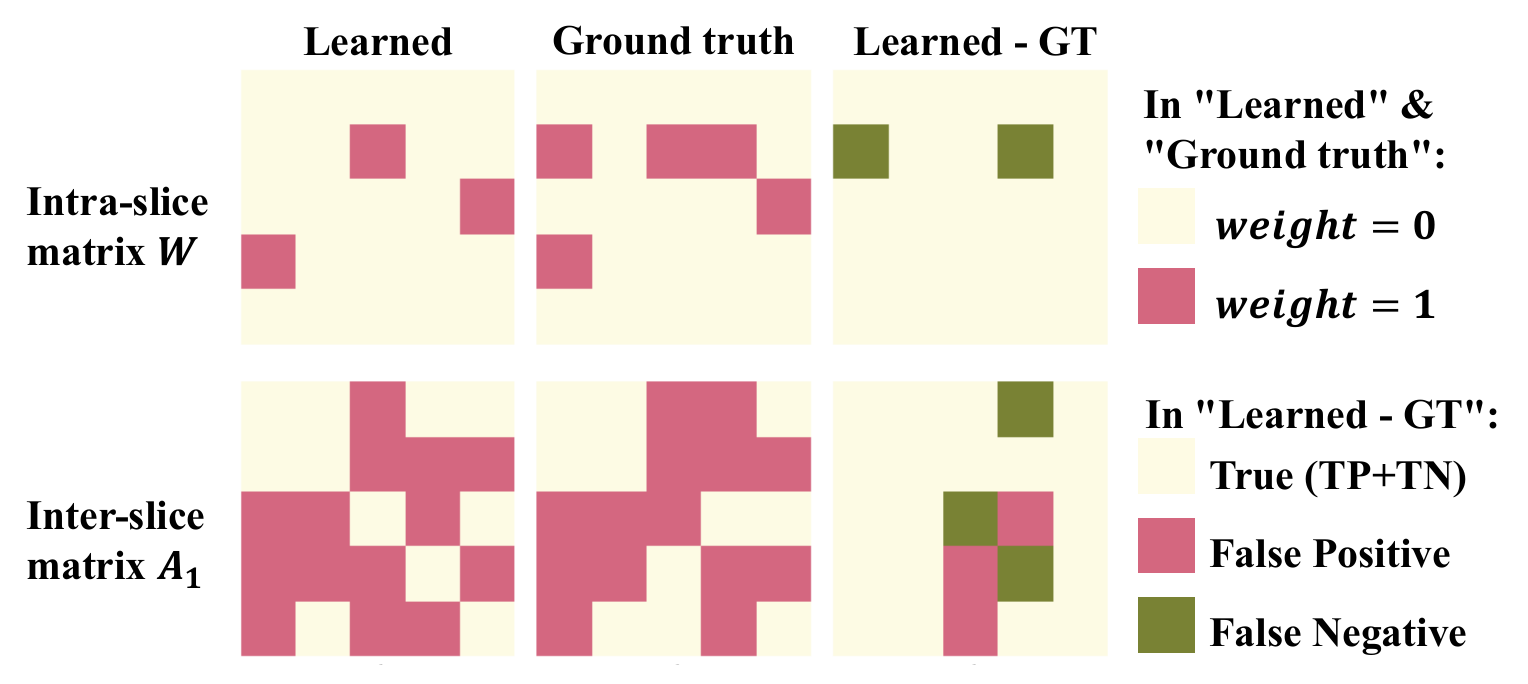}
    \caption{DAG results.}
    \label{fig:adj_simu}
  \end{subfigure}
  \begin{subfigure}[b]{0.7\linewidth}
    \centering
    \includegraphics[width=\linewidth]{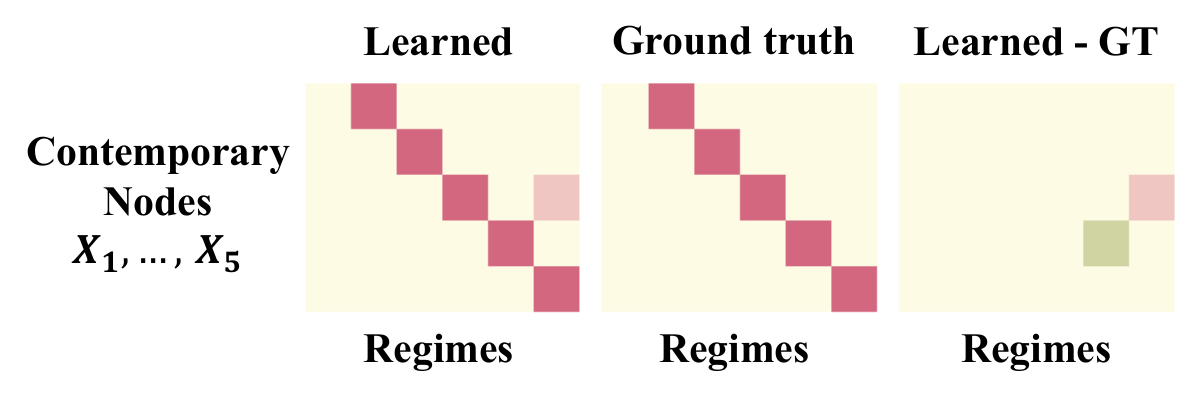}
    \caption{Interventional targets results.}
    \label{fig:target_simu}
  \end{subfigure}
  \caption{Showcases of the results on synthetic data.}
  \label{fig:combined}
\end{figure}

Figure~\ref{fig:adj_simu} and Figure~\ref{fig:target_simu} display example results of the temporal causal DAG and interventional targets obtained through \modelvnosp. 
In Figure~\ref{fig:target_simu}, each row represents a contemporary node, and each column represents a regime corresponding to each interventional family: the first for observational data and the next five for different interventional data. Pink cells indicate the learned or ground truth interventional targets in each regime. 
These figures demonstrate our method's ability to accurately identify both the temporal causal graph and interventional targets.


\subsubsection{On real-world data center datasets}
Table~\ref{tab:real} presents the results on real data. 
Our method achieves fewer C2A False and learns more A2A True and C2C True relations. Introducing prior domain knowledge with the LLM module effectively understands the upstream and downstream relationships in the system architecture, avoiding downstream influence on upstream variables and keeping C2A False at zero in the subsequent optimization. Additionally, the standard deviation of each metric is small. Therefore, \modelv outperforms other approaches in industrial scenarios by utilizing rich information from interventional data and prior domain knowledge from textual information.


\begin{table*}[h]
\small
\centering
\caption{Results on real-world datasets. $\uparrow$ denotes the higher the better, and $\downarrow$ denotes the lower the better. The best results are in bold in the setting of unknown interventional targets.}
\label{tab:real}
\begin{tabular}{lllccccc}
\toprule
\multicolumn{1}{l}{Targets} & Causality & Method & All edges & C2A False $\downarrow$ & A2C True $\uparrow$ & A2A True $\uparrow$ & C2C True $\uparrow$ \\
\midrule
\multirow{2}{*}{Known}&\multirow{2}{*}{intra+inter}& TECDI & \ms{85.60}{12.46}&\ms{4.00}{6.85}&\ms{5.80}{3.16}&\ms{0.70}{1.34}&\ms{2.80}{1.32}\\
 &  & \modelv & \ms{79.52}{7.96}&\msone{0.00}{0.00}&\msone{6.05}{1.26}&\msone{10.53}{2.15}&\msone{5.71}{3.04}\\
\cmidrule(r){1-8} 
\multirow{6}{*}{Unknown} & \multirow{4}{*}{intra+inter} & DYNOTEARS & \ms{38.00}{0.00}&\ms{0.00}{0.00}&\ms{0.00}{0.00} &\ms{0.00}{0.00}&\ms{0.00}{0.00}\\
 &  & PCMCI & \ms{181.60}{32.26}&\ms{32.90}{5.00}&\msone{13.20}{2.86}&\ms{7.50}{5.87}&\ms{5.00}{2.49}\\
 &  & TECDI & \ms{57.60}{10.67}&\ms{2.40}{1.51}&\ms{1.50}{1.84} &\ms{1.80}{1.48}&\ms{0.40}{0.97}\\
 &  & \modelv & \ms{51.70}{7.96}&\msone{0.00}{0.00}&\ms{1.50}{0.97}&\msone{14.00}{1.63}& \msone{8.40}{1.90}\\
 \cmidrule(r){2-8} 
     & \multirow{2}{*}{intra} & NeuralGC & \ms{104.10}{41.09}& \ms{25.70}{19.47}&\msone{1.90}{4.01} &\ms{5.20}{3.71}&\ms{4.00}{3.65}\\
 &  & \modelv  & \ms{51.70}{7.96}&\msone{0.00}{0.00}&\ms{1.50}{0.97}&\msone{12.00}{1.01}& \msone{6.80}{1.04}\\
 \bottomrule
\end{tabular}
\end{table*}

RealTCD in the "Unknown" targets setting, especially designed in our paper, achieves the best performance across all settings, even outperforming the "Known" targets setting. This is because the interventional targets provided by users are often not ground truth (people can only detect anomalous variables, but cannot confirm whether they are the ground truth interventional targets), which could potentially mislead the learning of algorithm. This further underscores the importance of using our method based on interventional data with unknown targets to deal with real-world cases.

\subsection{Deeper Analysis}

\subsubsection{Ablation studies}\label{Sec:abl}

We conduct ablation studies to verify the effectiveness of the proposed 2 modules in \model as in
Table~\ref{tab:ablreal}.

    
    

`Variant1' is `\modelv w/o intervention', which \textbf{removes the interventional module} and uses only observational data of the same sample size. It tends to output more edges, making comparisons unfair, so we calculate the ratio of labeled edges to total edges learned for further evaluation. Removing the interventional module significantly decreases performance, especially for the A2A True and C2C True ratios, confirming the effectiveness of our score-based temporal causal discovery from interventional data.

`Variant2' is `\modelv w/o LLM', which \textbf{removes the LLM-guided meta initialization}. Performance drops significantly without this module, even reaching zero for the A2A True metric. This shows that the LLM-guided meta initialization effectively utilizes prior information from system textual data, greatly enhancing temporal causal discovery in real-world scenarios. Integrating this module is crucial for providing a well-informed starting point, reducing biases, and improving causal inference precision.



\begin{table}[ht]
    \small
    \setlength{\columnwidth}{\linewidth}
    \centering
    \caption{Ablation studies of modules on real-world dataset.
    $\uparrow$ denotes the higher the better, and $\downarrow$ denotes the lower the better. The best results in terms of metric ratios are in bold.
    }
    \label{tab:ablreal}
    \begin{tabular}{cccccc}
    \toprule
     \footnotesize Method & \footnotesize All Edges & \footnotesize C2A False$\downarrow$ & \footnotesize A2C True$\uparrow$ & \footnotesize A2A True$\uparrow$ & \footnotesize C2C True$\uparrow$ \\
    \midrule
     \multirow{2}{*}{\footnotesize \modelvnosp}& \ms{51.7}{8.0} & \ms{0.0}{0.0} & \ms{1.5}{1.0} & \ms{14.0}{1.6} & \ms{8.4}{1.9} \\
      & \cellcolor{lightblue}\% & \cellcolor{lightblue}\ms{0.0}{0.0} & \cellcolor{lightblue}\ms{3.1}{2.4} & \cellcolor{lightblue}\msone{27.3}{2.7} & \cellcolor{lightblue}\msone{16.5}{3.6} \\
     \footnotesize \multirow{2}{*}{Variant1} & {{304.3}\scriptsize{}$\pm${18.7}} & \ms{0.0}{0.0} & \ms{43.3}{7.3} & \ms{16.5}{1.6} & \ms{14.8}{1.1} \\
     &\cellcolor{lightblue}\%  & \cellcolor{lightblue}\ms{0.0}{0.0} & \cellcolor{lightblue}\msone{14.2}{1.8} & \cellcolor{lightblue}\ms{5.4}{0.5} & \cellcolor{lightblue}\ms{4.9}{0.4} \\
     \footnotesize \multirow{2}{*}{Variant2} & \ms{38.8}{1.9} & \ms{0.0}{0.0} & \ms{0.1}{0.3} & \ms{0.0}{0.0} & \ms{0.4}{0.7} \\
     & \cellcolor{lightblue}\% & \cellcolor{lightblue}\ms{0.0}{0.0} & \cellcolor{lightblue}\ms{0.2}{0.7} & \cellcolor{lightblue}\ms{0.0}{0.0} & \cellcolor{lightblue}\ms{1.0}{1.6} \\
    \bottomrule
    \end{tabular}
\end{table}

\subsubsection{Different LLMs and prompts}
We tested different LLMs and prompts on real-world datasets~\cite{zhou2024causalbench,jin2024cladder}. GPT-4, especially with prompts containing implication and CoT, achieved better and more stable results. In scenarios where human ideas are unclear, zero-shot CoT can effectively leverage LLMs' domain knowledge, providing solutions that reveal the thinking process and enhance the interpretability of causal discovery.

\begin{table}[h]
\small
\caption{Comparisons of different LLMs and prompts. 
$\uparrow$ denotes the higher the better, and $\downarrow$ denotes the lower the better.
}
\label{tab:llm}
\centering
\setlength{\columnwidth}{\linewidth}
\begin{tabular}{ccccccc}
\toprule
Models & \multicolumn{3}{c}{GPT-4} & \multicolumn{3}{c}{GPT-3.5-turbo-instructor} \\
 \cmidrule(r){2-4} \cmidrule(r){5-7} 
Prompt No. &  0 & 1 & 2 &  0 & 1 & 2 \\ 
\texttt{Implication} & \checkmark & & & \checkmark & & \\
\texttt{CoT} & \checkmark & \checkmark & & \checkmark & \checkmark & \\
\midrule
C2A False $\downarrow$ & 0 & 0 & 0 & 0 & 0 & 0 \\
A2C True $\uparrow$ & 60 & \textbf{68} & 18 & 8 & 8 & 32\\
A2A True $\uparrow$ & \textbf{36} & 0 & 0 & 0 & 18 & 0\\
C2C True $\uparrow$ & \textbf{36} & 0 & 12 & 0 & 0 & 0\\
\bottomrule
\end{tabular}
\end{table}

\section{Discussion}


\subsection{Motivation and Strengths of Using LLMs}
In industrial systems, operations come with extensive textual documentation like logs and manuals, providing valuable insights into system behavior. LLMs excel at mining this textual data to inform causal analysis, directly connecting text and system operations.
We outline \textbf{the irreplaceable strengths of using LLMs} in \modelv over traditional deep learning approaches as follows:
\begin{itemize}[leftmargin = 0.5cm]
\item \textit{Handling Textual Information}: Traditional methods often ignore the rich textual data in real-world systems, while LLMs process and utilize it to enhance causal discovery accuracy. 
Though conventional language models can also process text, they lack the advanced capabilities of LLMs in in-context learning~\cite{cahyawijaya2024llms}, which provides LLMs with superior generalization and flexibility, as well as powerful zero-shot and few-shot learning abilities~\cite{brown2020language,li2024meta}.
\item \textit{Integration of Domain Knowledge}: LLMs assimilate user inputs  (e.g. the structure of a particular system) and integrate extensive domain-specific knowledge embedded within the LLMs (e.g. the operation law of a system), covering areas unfamiliar to users. This is crucial for accurate causal inference in complex systems.
\item \textit{LLM-guided Meta-Initialization}: This module significantly improves causal discovery quality by using meta-knowledge from LLMs to narrow the scope initially, unlike traditional methods that start with broad assumptions, which can lead to suboptimal local solutions and high variance in results.
\end{itemize}

The suitability of LLMs for enhancing causal discovery is also supported by a body of literature that demonstrates their effectiveness in extracting and applying domain knowledge in complex inference tasks~\cite{ban2023query,chen2023mitigating,kiciman2023causal,vashishtha2023causal,jiralerspong2024efficient}.

\subsection{Limitation}
We acknowledge that there are limitations related to the quality and completeness of the textual data in real-world applications~\cite{willig2023causal}. Issues such as noise in the data, incomplete or missing text, and small sample sizes can potentially affect the performance of our method. These challenges are indeed inevitable and deserve further investigation.

\subsection{Practical Implications}

In our paper, we highlight the application of temporal causal discovery within AIOps to enhance systems' monitoring, troubleshooting, and predictive capabilities, crucial for tasks such as anomaly detection~\cite{yang2022causality}, root cause analysis~\cite{wang2023incremental}, failure prediction, and system optimization. 
It is also helpful in various fields such as finance~\cite{kambadur2016temporal}, healthcare~\cite{gong2023causal,zhang2023ood,tong2024automating,naik2023applying}, and social sciences~\cite{grzymala2011time} by uncovering the dynamic interrelations between variables over time.
We discuss the \textbf{broad practical implications} elaborately in Appendix~\ref{App:imp}.

\section{Related Work}
\paragraph{Causal Discovery in Temporal Domain}
Interventional data greatly aids in identifying causal structures~\cite{eberhardt2012optimal,tian2013causal,gao2022idyno,zhang2023spectral}, but designing experiments and collecting data is challenging. ~\citet{jaber2020causal} used a $\Psi$-Markov property for learning causal graphs with latent variables. ~\cite{addanki2020efficient} proposed a $p$-collider-based algorithm to recover causal graphs with minimal intervention costs. ~\cite{brouillard2020differentiable} used interventional data and neural architectures for causality detection. \cite{li2023causal} focused on temporal causality from observational data but required intervention labels, limiting practicality. We focus on discovering temporal causal relationships without intervention labels.

\paragraph{LLMs for Causal Discovery}
Recent works explored LLMs for causal discovery\cite{jiralerspong2024efficient,Vashishtha2023CausalIU,Bhattacharjee2023TowardsLC,Gui2023TheCO,Kasetty2024EvaluatingIR,liu2024large,willig2022probing,takayama2024integrating,ban2023causal,cohrs2023large} and dynamic graphs\cite{zhang2023LLM4DyG,wang2017community,wang2019heterogeneous,zhang2023understanding}. \citet{ban2023query} showed LLMs' potential in causal inference. \citet{chan2023chatgpt} assessed ChatGPT's ability to capture temporal and causal relations. \citet{chen2023mitigating} used LLM-driven knowledge to reduce biases in causal learning. \cite{kiciman2023causal} highlighted LLMs' roles in causal discovery and reasoning. \citet{long2023can} investigated LLMs' understanding of causal relationships in medical contexts. 
To the best of our knowledge, we are the first to utilize LLMs in the field of temporal causal discovery.

\section{Conclusion}
In this paper, we propose the \model framework, a novel approach for temporal causal discovery in AIOps that bypasses the need for interventional targets and leverages system textual information. Our method, featuring score-based temporal causal discovery and LLM-guided meta-initialization, outperforms existing baselines on both simulated and real-world datasets. The results underscore \modelnosp's potential to enhance causal analysis in complex IT systems and suggest further research in AI-driven operations and maintenance. 


\begin{acks}
This work was supported in part by the National Key Research and Development Program of China No. 2020AAA0106300, National Natural Science Foundation of China (No. 62222209, 62250008, 62102222), Beijing National Research Center for Information Science and Technology Grant No. BNR2023RC01003, BNR2023TD03006, Alibaba Innovative Research Program and Beijing Key Lab of Networked Multimedia.
\end{acks}

\bibliographystyle{ACM-Reference-Format}
\bibliography{0_ref}

\newpage

\appendix
\section{Detailed Description of Datasets}\label{App:datasets}
\subsection{Synthetic Datasets Generation}
We generate synthetic temporal data in two distinct steps, meticulously designed to emulate realistic causal structures and dependencies:

\subsubsection{Graph structure sampling}
\begin{itemize}[leftmargin = 0.5cm]
    \item DAG Sampling: We first sample the structure of intra-slice and inter-slice Directed Acyclic Graphs (DAGs) following the \textit{Erdős-Rényi} model, a common approach for generating random graphs. This model allows us to control the sparsity of the DAG by setting a probability for the existence of each edge.
    \item Parameter Sampling: For the weighted adjacency matrices, we sample the edge weights to establish the strength and direction of causal influences:
        
        \textit{Intra-slice Matrix $W$}: The weights of the edges within the same time slice are sampled uniformly from the sets 
        $[-1.0,-0.25]\cup [0.25,1.0]$, ensuring non-trivial causal effects by excluding values close to zero.
        
        \textit{Inter-slice Matrices $A_k$}: For edges across different time slices, the weights are sampled uniformly from 
        $[-1.0\alpha,-0.25\alpha] \cup [0.25\alpha,1.0\alpha], \alpha=1/\eta^{k}, \eta\geq1, k=1,\ldots,p$. The parameter $k$ corresponds to the lag $k=1,\ldots,p$, introducing a decay factor that decreases the influence of past events over time, reflecting real-world temporal dynamics.
\end{itemize}

\subsubsection{Time series data generation}
\begin{itemize}[leftmargin = 0.5cm]
    \item Data Synthesis: Utilizing the sampled weighted graph, we generate time series data consistent with the standard structural vector autoregressive (SVAR) model as described by \citet{rubio2010structural}. The model equation $Y_0= Y_0 W+Y_1 A_1+\cdots+Y_p A_p+Z$ integrates both intra-slice and inter-slice dependencies, where $Y_0$ represents the current state variables, $Y_1,\ldots,Y_p$ represent past state variables up to lag $p$, and $Z$ is a vector of random variables drawn from a normal distribution, representing the noise in the system.
    \item Interventional Data: After generating the baseline time series data, we sample interventional targets from the nodes in $Y_0$. We then create perfect interventional scenarios by severing the dependencies of these targeted nodes from their respective parents. Specifically, we set $W_{ij}$ and ${A_k}_{ij}$ to zero, where $x_j$ is the variable in interventional targets and $x_i\in x_{\pi_j^\mathcal{G}}$.
\end{itemize}
By following this structured approach, we aim to generate a robust dataset that closely mimics the complexities and causal relationships inherent in real-world temporal systems.

\subsection{Real-world Data Center Datasets Generation}

Our LLM-enhanced temporal causal discovery method, which leverages interventional data, is \textbf{applicable across a wide range of industrial scenarios where both normal and anomalous states of data are present}. Examples include anomaly monitoring in power systems, analysis of factors affecting temperature, and financial risk monitoring. To date, there are no other type of publicly available real-world temporal datasets with interventional data suitable for evaluating such methods; previous studies typically rely on synthetic data. Consequently, we utilized the self-constructed dataset from a data center. However, we acknowledge the importance of diversifying our datasets and will seek to include more varied industrial data in future evaluations of our model.

\subsubsection{Scenario description}

In contemporary data centers, the stability of IT equipment operation is paramount. Sophisticated air conditioning systems play a critical role in maintaining the operational integrity of these facilities. They manage the heat generated by IT equipment, ensuring that the indoor temperature remains within safe operational limits. This thermal management is vital for preventing equipment malfunction or failure due to overheating.

The internal architecture of a standard data center room is methodically organized to optimize airflow and temperature control. As illustrated in Figure~\ref{fig:realgraph}, data centers are typically structured with rows of equipment cabinets labeled A, B, C, D, and so forth. These rows are separated by physical barriers that help isolate adjacent cabinet rows, effectively preventing the intermingling of hot and cold air streams. This isolation is crucial for efficient cooling as it helps maintain the integrity of cooled air distribution.

Strategically positioned along the sides of the room, Computer Room Air Conditioners (CRACs) play a pivotal role. These CRAC units facilitate a closed-loop configuration that is essential for preserving a stable environmental condition within the data center. By recycling the air within the room, CRACs ensure that the hot air exhausted from the server cabinets is cooled down and re-circulated.

Furthermore, the cooling architecture involves the use of chilled water systems, where cold air is not merely recycled but is generated through heat exchange with chilled water. The above system includes important components: \textit{Chilled Water Inlet}, which is used for heat exchange with the air; \textit{Water Valve Opening}, which controls the flow of chilled water; \textit{Return Air Temperature}, which measures the air temperature after it has absorbed heat from the IT equipment, indicating the efficiency of heat removal.

\textbf{Temperature sensors strategically deployed within the cold aisles collect real-time data.} This continuous monitoring allows for the dynamic adjustment of air conditioning settings to respond promptly to any fluctuations in temperature. Specifically, by controlling the temperature and flow of chilled water, the CRACs adjust the supply air temperature directed into the cold aisles.

This comprehensive and sophisticated thermal management system ensures that data centers can operate continuously without interruption, providing the necessary cooling to maintain IT equipment in optimal working conditions. This setup not only protects valuable IT assets but also enhances energy efficiency and reduces the risk of downtime caused by equipment overheating.

\subsubsection{Data acquisition}

The dataset used in this study was sourced from a specific data center at Alibaba. It primarily focuses on the subsystem illustrated in the upper half of Figure~\ref{fig:realgraph}, capturing the relationships between \textbf{air conditioning supply temperatures} and \textbf{cold aisle temperatures}. Specifically, it includes monitoring data from the cooling system of a particular room, covering the period from January 1st, 2023, to May 1st, 2023. In total, the dataset comprises 38 variables, which consist of 18 cold aisle temperatures from sensors located in the cold aisles and 20 air conditioning supply temperatures from Computer Room Air Conditioners (CRACs).

Due to strict alarm controls in the data center, overall alarms are quite rare. Even though some sensors exhibit fluctuations, these are typically controlled and do not trigger alarms. Hence, the data provided does not represent real alarm data but rather "anomalies" in the supply air temperatures and cold aisle temperatures. Anomaly points were identified by learning the normal distribution range from historical data, using the $n$-$\sigma$ method. Any data points that fell outside the $n$-$\sigma$ range (e.g., 3 to 5 times the standard deviation) of their respective variables were extracted as anomaly time points.

To collect this data, we acquired several time series of these 38 variables during both normal and abnormal states. During the abnormal conditions, data was sampled within 20 minutes of the detection of an anomaly, with each sampling interval set at 10 seconds.

\subsubsection{Detailed data acquisition method}
\begin{itemize}[leftmargin=0.5cm]
    \item \textbf{Data Collection Scope}: The dataset encapsulates temperature readings from both the air conditioning supply (CRACs) and the cold aisles in a specific room of an Alibaba data center.
    \item \textbf{Time Frame}: Data was continuously monitored and recorded from January 1st, 2023, to May 1st, 2023.
    \item \textbf{Variable Details}: The dataset includes 38 distinct temperature variables, with 18 sourced from sensors in the cold aisles and 20 from the air conditioning supply units.
    \item \textbf{Anomaly Detection}: To identify anomalies in temperature readings, historical data was analyzed to establish a normal distribution range for each variable. Using the $n$-$\sigma$ method, any readings that deviated significantly (3 to 5 times the standard deviation) from this norm were flagged as anomalies.
    \item \textbf{Data Sampling for Anomalies}: For anomalies identified, additional data was collected at a high resolution. Data points were sampled every 10 seconds within a 20-minute window surrounding each identified anomaly, ensuring a detailed capture of conditions during abnormal states.
\end{itemize}

This approach allows for a comprehensive analysis of the cooling system's performance and the detection of any deviations from expected operational conditions. By modeling the anomalies into interventional data and further understanding these anomalies, the data center can optimize its cooling strategy and improve the overall stability of the IT equipment environment.

\section{Example Prompt}\label{App:prompts}
We provide a specific example of the prompt used for the data center dataset in our experiments, shown in Table~\ref{tab:detailprop}.

\begin{table*}
    \caption{Prompts example for data center datasets.}
    \begin{tabular}{l|p{15cm}}
    \toprule
        \textbf{Prompt D}& 
        \textbf{Role}: You are an exceptional temporal causal discovery analyzer, with in-depth domain knowledge in the intelligent operation and maintenance of data center air-conditioning systems. \\
         & \\
         & \textbf{Introduction}: A directed temporal causal relationship between variables xu and xv can be represented as a tuple (xu, xv, t), signifying that the variable xu, lagging t time units, causally influences the current state of variable xv. The tuple (xu, xv, 0) denotes contemporaneous causality if t=0; if t>0, the tuple (xu, xv, t) indicates time-lagged causality. Note that xu and xv must be different variables, as self-causality is not considered. Also, (xu, xv, 0) and (xu, xv, t) for t>0 have the possibility to coexist, suggesting that contemporaneous and time-lagged causality between two variables might simultaneously occur sometimes.\\
         & \\
         & Our task is to unearth temporal causal relationships among variables, grounded on the subsequent information. \\
         & \\
         \textbf{Prompt I}&
         \textbf{Domain knowledge}: In modern data centers, maintaining stable IT equipment operation is critical. Advanced air conditioning systems manage the heat produced by equipment, sustaining a stable indoor climate. Equipment rows in a typical data center room are methodically arranged, with physical barriers separating adjacent rows to prevent air mixing. Computer room air conditioners (CRACs) flank the room, circulating cooling air and forming a closed loop for environmental stability. Multiple sensors in the cold aisle continuously monitor temperature, enabling prompt adjustments for stable IT equipment operation. \\
         & \\
         & There are 38 measuring points (20 CRACs and 18 sensors), represented by ‘measuring{\_}points{\_}name:(x-axis{\_}value, y-axis{\_}value)' to indicate their relative positions. Using the bottom-left corner (CRAC{\_}11) as the coordinate origin (0,0) and meters as the unit, the positions are as follows. There are 10 CRACs in the top row, from left to right: CRAC{\_}1:(0,10.8), CRAC{\_}2:(2.4,10.8), CRAC{\_}3:(3.4,10.8), CRAC{\_}4:(5.4,10.8), CRAC{\_}5:(6.6,10.8), CRAC{\_}6:(9,10.8), CRAC{\_}7:(10.2,10.8), CRAC{\_}8:(12.4,10.8), CRAC{\_}9:(13.6,10.8), CRAC{\_}10:(17,10.8); also 10 CRACs in the bottom row, from left to right: CRAC{\_}11:(0,0), CRAC{\_}12:(1.8,0), CRAC{\_}13:(3.6,0), CRAC{\_}14:(5.4,0), CRAC{\_}15:(6.6,0), CRAC{\_}16:(8.4,0), CRAC{\_}17:(10.8,0), CRAC{\_}18:(13.2,0), CRAC{\_}19:(14.4,0), CRAC{\_}20:(17,0). There are 5 columns of cabinets in the middle of the room, so there are 6 columns cold channels separated, each with three sensors, from the bottom right corner to the top left corner in turn: sensor{\_}1:(17,3.9), sensor{\_}2:(17,5.4), sensor{\_}3:(17,7.5), sensor{\_}4:(13.6,3.9), sensor{\_}5:(13.6,5.4), sensor{\_}6:(13.6,7.5), sensor{\_}7:(10.2,3.9), sensor{\_}8:(10.2,5.4), sensor{\_}9:(10.2,7.5), sensor{\_}10:(6.8,3.9), sensor{\_}11:(6.8,5.4), sensor{\_}12:(6.8,7.5), sensor{\_}13:(3.4,3.9), sensor{\_}14:(3.4,5.4), sensor{\_}15:(3.4,7.5), sensor{\_}16:(0,3.9), sensor{\_}17:(0,5.4), sensor{\_}18:(0,7.5). 
        From the coordinates provided, you can know the relative position of the measuring points of the various devices in the computer room.\\
        & \\
        & Correspondingly, we consider 38 variables: air{\_}1, ..., air{\_}20 representing the supply temperatures of CRAC{\_}1, ..., CRAC{\_}20, and cold{\_}1, ..., cold{\_}18 denoting the temperatures in the cold aisle monitored by sensor{\_}1, ..., sensor{\_}18. \\
        & \\
        \textbf{Prompt C}&
        \textbf{Task}: Please identify all temporal causal relationships among the 38 variables (air{\_}1 to air{\_}20 and cold{\_}1 to cold{\_}18), considering only contemporaneous and 1 time-lagged causality (t=0 or 1). Conclude your response with the full answer as a Python list of tuples (xu, xv, t) after 'Answer:'. Write complete answers, don't cut them down.\\
        & \\
        \textbf{Prompt H}& 
        \textbf{CoT}: Proceed methodically, step by step.\\
    \bottomrule
    \end{tabular}
    \label{tab:detailprop}
\end{table*}

\section{Practical Implications}\label{App:imp}
In our paper, we emphasize the application of temporal causal discovery within AIOps to enhance IT systems' monitoring, troubleshooting, and predictive capabilities. This approach is crucial for improving system monitoring, anomaly detection[1], root cause analysis~\cite{wang2023incremental}, failure prediction, and overall system optimization by elucidating the dynamic interactions between different system components and events over time.

Moreover, the utility of temporal causal discovery extends beyond IT operations into several other domains, thereby broadening our understanding and ability to manipulate complex systems across various fields:
\begin{itemize}[leftmargin=0.5cm]
    \item Finance and Economics: Temporal causal models can refine financial forecasting and investment strategies by revealing how economic variables influence each other over time, leading to more informed and effective decision-making~\cite{kambadur2016temporal}.
    \item Healthcare: In medical research, understanding the temporal sequences of cause and effect can aid in predicting disease progression, optimizing treatment plans, and improving patient outcomes~\cite{gong2023causal,zhang2023ood,tong2024automating,naik2023applying}.
    \item Social Sciences and Governance: Temporal analysis is instrumental in political science and history for identifying causative factors behind regime changes and policy impacts, thus enhancing theoretical frameworks for understanding institutional transformations~\cite{grzymala2011time}.
\end{itemize}

These examples illustrate the broad applicability and significant impact of temporal causal discovery in various sectors.

\end{document}